\pdfoutput=1
\documentclass[nohyperref]{article}

\usepackage{microtype}
\usepackage{graphicx}
\usepackage{subfigure}
\usepackage{booktabs} 
\usepackage{multirow}

\usepackage{hyperref}

\usepackage[accepted]{icml2022}

\usepackage{amsmath}
\usepackage{amssymb}
\usepackage{mathtools}
\usepackage{amsthm}

\usepackage{pgfplots}
\usepackage{stfloats}

\definecolor{transfertoserver}{HTML}{D7191C}
\definecolor{database}{HTML}{FDAE61}
\definecolor{transfertoclient}{HTML}{ABDDA4}
\definecolor{rendering}{HTML}{2B83BA}
\definecolor{foo}{HTML}{EFF5F9}
\definecolor{magenta}{HTML}{FF00FF}

\usepackage[capitalize,noabbrev]{cleveref}

\theoremstyle{plain}

\theoremstyle{definition}

\theoremstyle{remark}

\usepackage[textsize=tiny]{todonotes}

\usepackage{placeins}

\icmltitlerunning{The Combinatorial Brain Surgeon: Pruning Weights That Cancel One Another in Neural Networks}

\usepackage{enumerate}

\usepackage{amsmath,amsfonts,bm}

\def\eqref#1{equation~\ref{#1}}

\def\1{\bm{1}}

\def\ve{{\bm{e}}}

\def\vw{{\bm{w}}}

\def\vy{{\bm{y}}}

\def\evw{{w}}

\def\mA{{\bm{A}}}

\def\mH{{\bm{H}}}

\DeclareMathAlphabet{\mathsfit}{\encodingdefault}{\sfdefault}{m}{sl}
\SetMathAlphabet{\mathsfit}{bold}{\encodingdefault}{\sfdefault}{bx}{n}

\def\sB{{\mathbb{B}}}

\def\sI{{\mathbb{I}}}
\def\sJ{{\mathbb{J}}}

\def\sP{{\mathbb{P}}}
\def\sQ{{\mathbb{Q}}}

\def\sS{{\mathbb{S}}}

\def\sW{{\mathbb{W}}}

\newcommand{\best}[1]{\textcolor{blue}{\bf #1}}
\newcommand{\second}[1]{\textcolor{orange}{\bf #1}}

\usepackage[symbol]{footmisc}

\newcommand{\alglinelabel}{%
  \addtocounter{ALC@line}{-1}
  \refstepcounter{ALC@line}
  \label
}

\usepackage{placeins}

\usepackage{epigraph} 

\definecolor{bblue}{rgb}{0,0.15,0.25}
\definecolor{bmblue}{rgb}{0,0.51,0.73}
\definecolor{borange}{rgb}{0.97,0.3,0.09}
\definecolor{cred}{rgb}{0.6,0,0}

\begin{document}

\twocolumn[
\icmltitle{The Combinatorial Brain Surgeon: \\ Pruning Weights That Cancel One Another in Neural Networks}



\icmlsetsymbol{equal}{*}

\begin{icmlauthorlist}
\icmlauthor{Xin Yu}{equal,uu}
\icmlauthor{Thiago Serra}{equal,bu}
\icmlauthor{Srikumar Ramalingam}{gr}
\icmlauthor{Shandian Zhe}{uu}
\end{icmlauthorlist}

\icmlaffiliation{uu}{University of Utah, Salt Lake City, UT, United States}
\icmlaffiliation{bu}{Bucknell University, Lewisburg, PA, United States}
\icmlaffiliation{gr}{Google Research, New York, NY, United States}

\icmlcorrespondingauthor{Shandian Zhe}{zhe@cs.utah.edu}
\icmlcorrespondingauthor{Thiago Serra}{thiago.serra@bucknell.edu}

\icmlkeywords{Machine Learning, ICML}

\vskip 0.3in
]



\printAffiliationsAndNotice{\icmlEqualContribution} 

\begin{abstract}
Neural networks tend to achieve better accuracy with training if they are larger --- even if the resulting models are overparameterized. 
Nevertheless, carefully removing such excess of parameters before, during, or after training may also produce models with similar or even improved accuracy.  
In many cases, that can be curiously achieved by heuristics as simple as removing a percentage of the weights with the smallest absolute value --- even though absolute value is not a perfect proxy for weight relevance. 
With the premise that obtaining significantly better performance from pruning depends on accounting for the combined effect of removing multiple weights, 
we revisit one of the classic approaches for impact-based pruning: the Optimal Brain Surgeon~(OBS). 
We propose a tractable heuristic for solving the combinatorial extension of OBS, in which we select weights for simultaneous removal, and we combine it with a single-pass systematic update of unpruned weights.  
Our selection method outperforms other methods for high sparsity, and the single-pass weight update is also advantageous if applied  after those methods. 
\textbf{Source code:}   
\url{github.com/yuxwind/CBS}.
\end{abstract}

\section{Introduction}

In a world where large and overparameterized neural networks keep GPUs burning hot  
because machine learning researchers rethought generalization~\citep{zhang2017rethinking,belkin2019nas} 
and started tuning neural architectures with a hope for globally convergent loss landscapes~\citep{li2018landscape,ruoyu2020landscape}, 
network pruning can perhaps save us from parameter redundancy~\citep{denil2013parameters}.

Network pruning can lead to more parameter-efficient networks, with which we can save on model deploying and storage costs, and even to models with better accuracy. 
Although the extent to which we can prune depends on the task~\citep{liebenwein2021lost} 
and pruning may have an uneven impact across classes if not done properly~\citep{hooker2019forget,paganini2020responsibly,hooker2020bias,good2022recall}, 
pruning can also make neural networks more robust to adversarial manipulation \citep{wu2021adversarial}. 

From a perspective of model expressiveness using linear regions, 
we may see network pruning as a means to close the gap between the highly complex models that can be theoretically learned with an architecture~\citep{pascanu2013on,montufar2014on,Telgarsky2015,montufar2017notes,arora2018understanding,serra2018bounding,serra2020empirical,xiong2020cnn,montufar2021maxout} 
and the relatively less complex models obtained in practice~\citep{hanin2019complexity,hanin2019deep,tseran2021expected}. 

Curiously, however, the long-standing magnitude-based pruning approach~\citep{hanson1988minimal,mozer1989relevance,janowsky1989prunning}
remains remarkably competitive~\citep{blalock2020survey} despite having equally long-standing evidence that it does not offer a good proxy for parameter relevance~\citep{hassibi1992surgeon}. But why?

We conjecture that focusing on the impact of removing each parameter alone prevents impact-based methods from being more effective. 
In other words, 
we believe that the decision about pruning each parameter should be based on which other parameters are also removed. 
Ideally, we want the effect of such removals to cancel one another to the largest extent possible while achieving the aimed sparsity. 

In this work, 
we revisit 
the functional Taylor expansion of the loss function $\mathcal{L}$ on a choice of weights $\vw \in \mathbb{R}^N$ around the learned weights $\bar{\vw} \in \mathbb{R}^N$ of the neural network as 
\begin{align*}
\mathcal{L}(\vw) - \mathcal{L}(\bar{\vw}) = & (\vw - \bar{\vw})^T\nabla \mathcal{L}(\bar{\vw})  \\ & + \frac{1}{2}(\vw - \bar{\vw})^T \nabla^2 \mathcal{L} (\bar{\vw}) (\vw - \bar{\vw}) \\ & + O(\| \vw - \bar{\vw} \|^3),
\end{align*}
which is the basis for classic methods 
such as Optimal Brain Damage~(OBD) by
\citet{lecun1989damage} and Optimal Brain Surgeon~(OBS) by \citet{hassibi1992surgeon}. 

Like in OBD and OBS, we assume that (i) the training converged to a local minimum, so $\nabla \mathcal{L}(\bar{\vw}) = 0$; and (ii) $\vw$ is sufficiently close to $\bar{\vw}$, so $O(\| \vw - \bar{\vw} \|^3) \approx 0$. Hence, 
\begin{align}
\mathcal{L}(\vw) - \mathcal{L}(\bar{\vw}) \approx  \frac{1}{2} (\vw - \bar{\vw})^T \nabla^2 \mathcal{L} (\bar{\vw}) (\vw - \bar{\vw}).
\end{align}
Under such conditions, we can assess the impact of pruning the $i$-th weight with $\evw_i = 0$ and $\evw_j = \bar{\evw}_j ~ \forall j \neq i$.

Local quadratic models of the loss function have been used in varied ways, then and now. 
In OBD, the Hessian matrix $\mH := \nabla^2 \mathcal{L} (\tilde{\vw})$ is further assumed to be diagonal, 
hence implying that pruning one weight has no impact on pruning the remaining weights. 
In OBS, the Hessian matrix $\mH$ is no longer assumed to be diagonal, and the optimality conditions are used to update the remaining weights of the network based on removing the weight which causes the least approximate increase to the loss function. 
The first modern revival of this approach is the Layerwise OBS by \citet{dong2017layerwise}, 
in which each layer is pruned independently from the others. 
More recently, 
WoodFisher by \citet{singh2020woodfisher} operates over all the layers by introducing a new method that more efficiently approximates the Hessian inverse $\mH^{-1}$, which is necessary for the weight updates of OBS. 
In a sense, those modern revivals focused on keeping the calculation of $\mH^{-1}$ manageable. 

However, 
current OBS approaches do not account for the effect of one pruning decision on other pruning decisions. 
Whereas OBS does take into account the impact of pruning a given weight on the unpruned weights, only the weight with smallest approximated impact on the loss function is pruned at each step. 
In the most recent work, 
\citet{singh2020woodfisher} described the updates involved on choosing two weights for simultaneous removal, 
but nevertheless observed that considering the removal of multiple weights together would be impractical if performed in such a way.

While not disagreeing with \citeauthor{singh2020woodfisher}'s stance, 
we nevertheless proceed to formulate this problem and then consider how to approach it in a tractable way. In a nutshell, 
the contributions of this paper are the following:
\begin{enumerate}[(i)]
    \item We propose a formulation of the Combinatorial Brain Surgeon~(CBS) problem using Mixed-Integer Quadratic Programming~(MIQP), which for tractability is decomposed into the problems of pruned weight selection and unpruned weight update  (Section~\ref{sec:cbs});
    \item We propose a local search algorithm to improve the selection of the weights to be pruned  (Section~\ref{sec:local_search});
    \item We propose a randomized extension of magnitude-based pruning to obtain a diverse pool of starting points for the local search algorithm  (Section~\ref{sec:rmp}); and
    \item We decouple the systematic weight update from weight selection to circumvent the scalability issue anticipated by \citet{singh2020woodfisher} (Section~\ref{sec:update}).
\end{enumerate}

We focus on single-shot pruning after training to make before and after comparisons easier;  as well as to facilitate comparing the results of our method with existing work, 
in particular that of \citet{singh2020woodfisher}. 
We discuss additional related work in Section~\ref{sec:literature}, evaluate the proposed algorithms in Section~\ref{sec:experiments}, and draw conclusions in Section~\ref{sec:conclusion}.

\section{Related Work}\label{sec:literature}

\citet{blalock2020survey} observes that most work in network pruning relies on either magnitude-based or impact-based methods for selecting which weights to remove from the neural network. 
In fact, the list of key references for each is so long that we will use different paragraphs for each.

Magnitude-based methods select the weights with smallest absolute value~\citep{hanson1988minimal,mozer1989relevance,janowsky1989prunning,han2015connections,han2016deepcompression,li2017cnn,frankle2019lottery,elesedy2020lth,gordon2020bert,tanaka2020synapticflow,liu2021sparse}. 

Impact-based methods aim to select weights which would have less impact on the model if removed. That includes gradient-based approaches such as ours but we also regard other approaches~\cite{lecun1989damage,hassibi1992surgeon,hassibi1993surgeon,lebedev2016braindamage,molchanov2017taylor,dong2017layerwise,yu2018importance,zeng2018mlprune,baykal2019coresets,lee2019pretraining,wang2019eigendamage,liebenwein2020provable,wang2020tickets,xing2020lossless,singh2020woodfisher}. 

To those we can add the recent stream of exact methods, which aim to preserve the model intact while pruning the network~\citep{serra2020lossless,sourek2021lossless,serra2021compression,chen2021once,iclr2022compression}. 
To the best of our understanding, these methods are currently only beneficial under specific conditions. 

In addition to the discussion of pruning two parameters under OBS by \citet{singh2020woodfisher}, 
other recent works consider joint parameter pruning in neural networks.   
\citet{chen2021once} partition the parameters into zero-invariant groups for removal. 
\citet{liu2021group} identify coupled channels that should be either kept or pruned together. 

Across all these types of approaches, most work has been done on pruning trained neural networks and then fine tuning them afterwards. However, there is an growing body of work on pruning during training or at initialization~\cite{frankle2019lottery,lee2019pretraining,liu2019structure,lee2020pretraining,wang2020tickets,renda2020retraining,tanaka2020synapticflow,frankle2021initialization,zhang2021lessdata}.

The interest for the latter topic is in part attributed to the Lottery Ticket Hypothesis (LTH), 
according to which randomly initialized dense neural networks contain subnetworks that can be trained to achieve similar accuracy as the original network~\citep{frankle2019lottery}. 
That also led to work on the strong LTH, according to which one can improve the accuracy of randomly initialized models by pruning instead of training~\citep{zhou2019lth,ramanujan2020,malach2020lth,pensia2020lth,orseau2020lth,qian2021lth,daiki2021randomization}.

Another common theme is the formulation of optimization models for network pruning, which include other references not mentioned above~\citep{he2017cnn,luo2017thinet,aghasi2017nettrim,elaraby2020importance,ye2020forward,verma2021subdifferential,ebrahimi2021kfac}.

There are also more general approaches that would be better described as compression than as pruning methods, 
such as combining neurons and 
low-rank approximation, factorization, and random projection of weight matrices~\cite{jaderberg2014lowrank,denton2014linear,lebedev2015decomposition,srinivas2015datafree,mariet2016diversity,arora2018bounds,wang2018wide,su2018tensorial,wang2019eigendamage,suzuki2020spectral,suzuki2020bounds,suau2020distillation,li2020understanding}.

\section{Pruning as an Optimization Problem: the Combinatorial Brain Surgeon}\label{sec:cbs}

We formulate the Combinatorial Brain Surgeon~(CBS) as an optimization problem based on the local quadratic model of the loss function previously described.  

For a sparsity rate $r \in (0,1]$ corresponding to the fraction of the weights $\vw \in \mathbb{R}^N$ of a neural network to be pruned, 
we formulate the problem of selecting the $\lceil r N \rceil$ weights to remove and the remaining $N - \lceil r N \rceil$ weights to update with minimum approximate loss. 
In order to model the loss, we use $\bar{\evw}$ to denote the weights of the trained neural network before pruning, as well as $\mH$ to denote the Hessian matrix $\nabla^2 \mathcal{L}(\bar{\vw})$ and $\mH_{i,j}$ to denote the element at the $i$-th row and $j$-th column. 
That yields the following Mixed-Integer Quadratic Programming~(MIQP) formulation:
\begin{align}
    \min ~~~~ & \frac{1}{2} \sum_{i=1}^N \sum_{j=1}^N (\evw_i - \bar{\evw}_i) \mH_{i,j} (\evw_j - \bar{\evw}_j) \\
    \text{subject to} ~~~~ & \sum_{i=1}^N y_i = \lceil r N \rceil \\
    & y_i \rightarrow \evw_i = 0 \qquad \qquad \forall i \in \{1, \ldots, N \} \label{eq:indicator} \\
    & y_i \in \{0, 1\} ~~~ \qquad \qquad \forall i \in \{1, \ldots, N \} \\
    & \evw_i \in \mathbb{R} ~~~~~~~~~ \qquad \qquad \forall i \in \{1, \ldots, N \}
\end{align}
The decision variables of this formulation are, for each weight $i$, the updated value $\evw_i$ of that weight and a binary variable $y_i$ denoting whether the weight is pruned or not.

In order to obtain a tractable approach to CBS, we consider two special cases of this formulation in what follows.

\paragraph{CBS Selection} 
First we consider the CBS Selection~(CBS-S) formulation, in which we aim to minimize the approximate loss from pruning a selection of weights without updating the unpruned weights. For the formulation above, 
that means $\evw_i = \bar{\evw}_i$ if $y_i = 0$ and $\evw_i = 0$ if $y_i=1$. 
With each weight having a binary domain, $\evw_i \in \{0, \bar{\evw}_i\}$, only pairs of weights which are both removed affect the objective function. Hence, we can abstract the decision variables associated with weights and avoid implementing the indicator constraint in~\cref{eq:indicator}, which could potentially lead to numerical difficulties. That yields  the following Integer Quadratic Programming~(IQP) formulation:
\begin{align}
    \min ~~~~ & \frac{1}{2} \sum_{i=1}^N \sum_{j=1}^N \bar{\evw}_i y_i \mH_{i,j} \bar{\evw}_j  y_j \label{eq:cbs-u}\\
    \text{subject to} ~~~~ & \sum_{i=1}^N y_i = \lceil r N \rceil \\
    & y_i \in \{0, 1\} ~~~ \qquad \qquad \forall i \in \{1, \ldots, N \}
\end{align}

The formulation above is at the core of how we identify a combination of pruned weights affecting the local approximation of the loss function by the least amount. 
Namely, the impact of pruning both weights $i$ and $j$ is captured by $\frac{1}{2} \left( \evw_i \mH_{i,j} \evw_j + \evw_j \mH_{j,i} \evw_i \right)$\footnote[2]{For generality, we do not assume $\mH$ to be symmetric.}. 
In other words, the impact of pruning a weight depends on the other pruned weights. 
If weight updates are not considered, CBS-S is all you need. 

We can linearize this formulation by replacing $y_i y_i$ with $y_i$ and replacing $y_i y_j, i \neq j,$ with a binary decision variable $z_{i,j}$ and the constraints $z_{i,j} \leq y_i$, $z_{i,j} \leq y_j$, and $z_{i,j} \geq y_i + y_j - 1$ \citep{padberg1989}. However, that would make the number of variables and constraints grow quadratically, which would be impractical for large neural networks. 

\paragraph{CBS Update} 
Next we consider the CBS Update (CBS-U) formulation, in which we aim to minimize the approximate loss from updating the unpruned weights. 
Given a solution $\tilde{\vy}$ of CBS-S, we state that $\evw_i = 0$ if $\tilde{y}_i=0$. 
That yields the following Quadratic Programming~(QP) formulation:  
\begin{align}
    \min ~~~~ & \frac{1}{2} \sum_{i=1}^N \sum_{j=1}^N (\evw_i - \bar{\evw}_i) \mH_{i,j} (\evw_j - \bar{\evw}_j) \\
    \text{subject to} ~~~~ 
    & \evw_i = 0 \qquad ~~ \forall i \in \{1, \ldots, N \} : \tilde{y}_i = 1 \\
    & \evw_i \in \mathbb{R}  \qquad ~~ \forall i \in \{1, \ldots, N \} : \tilde{y}_i = 0
\end{align}
By abstracting the pruned weights altogether, we can reformulate CBS-U as an unconstrained quadratic optimization problem which can be efficiently solved (see Section~\ref{sec:update}).

Dissociating CBS into two subproblems has consequences, good and bad. 
On the one hand, 
solving each subproblem to optimality does not necessarily imply that we would obtain an optimal solution to CBS. 
However, 
it is very unlikely that we would obtain an optimal solution even to CBS-S for neural networks of reasonable size in the first place. 
If we were to nevertheless contemplate such a possibility, 
we could use a Benders-type decomposition~\citep{benders1962,hooker2003lbbd} through which we would alternate between solving a variant of CBS-S and CBS-U as formulated above by iteratively adding additional constraints to the initial formulation of CBS-S. 
These constraints would capture the change to the loss function due to weight updates for each selection of weights to prune obtained by solving the variant of CBS-S in prior steps. 

On the other hand, 
this dissociation of the weight selection and weight update problems leaves most of the computational difficulty to the former. 
This is particularly beneficial because it allows us to explore tried-and-tested techniques commonly used to solve discrete optimization problems.

Although CBS-S --- and even CBS --- could potentially be fed into Mixed-Integer Programming~(MIP) solvers capable of producing optimal solutions, 
that would not scale to neural networks of reasonable sizes.  
We refer the reader interested in formulations and algorithms for MIP problems to \citet{ccz2014book}. 
Interestingly, 
the algorithmic improvements of 
MIP solvers have been comparable to their contemporary hardware improvements for decades~\citep{bixby2012history}, 
but before that happened the tricks of the trade were  different: they involved designing good heuristics. 
In the next sections, we will resort to heuristic techniques that allowed optimizers to obtain good solutions for seemingly intractable problems subject to the solvers and hardware of their time --- and even to those of today in cases like ours.

\section{A Greedy Swapping Local Search Algorithm to Improve Pruning Selection}\label{sec:local_search}

The local search is where we take full advantage of the interdependence between pruned weights in our approach. 
Local search methods are used to produce better solutions --- or a diversified pool of solutions --- based on adjusting an initial solution, 
which in our case would be a selection of weights $\sP$ to be pruned. 
We refer the reader interested in local search to 
\citet{ls1996book}. 

In particular, 
our local search method iteratively swaps a weight in $\sP$ with a weight not in $\sP$. 
Similar operations have been long used for the traveling salesperson problem~\citep{tsp2006book}, 
in which the 2-opt~\citep{flood1956tsp,croes1958inversion} and the 3-opt~\citep{lin1965tsp} algorithms 
respectively swap two and three arcs from the tour 
with other arcs having a smaller sum of weights 
as long as an improvement is found. 
In our case, 
since minimizing the loss function on the training set may not necessarily lead to better test set accuracy, 
we found that avoiding swaps with negligible loss improvement led to better results. 

\begin{figure}[h]
\centering
\includegraphics[width=0.48\textwidth]{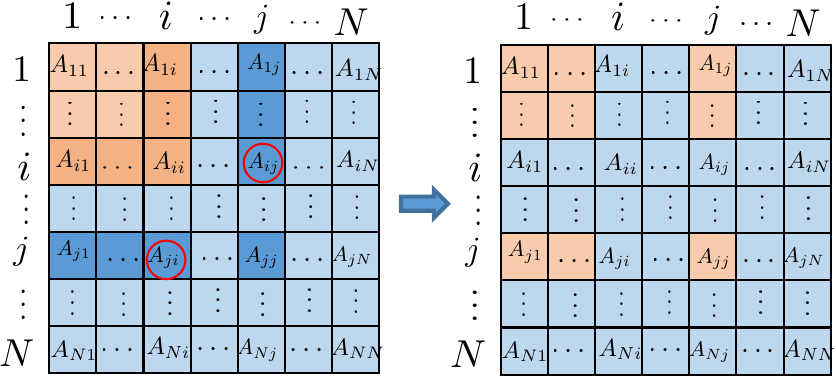}
\caption{Swapping weight $i\in \mathcal{\sP}$ and $j \in \mathcal{\bar{\sP}}$ in Algorithm~\ref{alg:local_search}. Assume that the first $i$ weights are in $\sP$ and others are in $\bar\sP$ before pruning (left). Each $\mA_{km} \coloneqq \bar{\evw}_k \mH_{k,m} \bar{\evw}_m$. Note that only when both weights $k$ and $m$ are selected, namely $y_k=y_m=1$, $\mA_{km}$ counts in the objective as in~\eqref{eq:cbs-u}. We highlight such elements in \textcolor{Tan}{orange} in both left and right sub-figures for before and after swapping respectively. 
We further highlight the elements used by \textcolor{orange}{$\boldsymbol\alpha_i$} in \textcolor{orange}{dark orange}, elements used by \textcolor{RoyalBlue}{$\boldsymbol\beta_j$} in \textcolor{RoyalBlue}{dark blue}, elements used by \textcolor{BrickRed}{$\boldsymbol\gamma_{ij}$} with \textcolor{BrickRed}{red circles} before swapping (left). The objective change after pruning is $-\alpha_i + \beta_j - \gamma_{i,j}$.}
\label{fig:vis-cbs-s}
\end{figure}

\cref{alg:local_search} describes our local search method. 
The outer loop repeats for $steps_{max}$ steps, unless the sample loss is not improved in $noimp_{max}$ consecutive steps or a step concludes without changing the set of pruned weights $\sP$. 
At every repetition of the outer loop, 
we initialize the sets $\sI \subseteq \sP$ and $\sJ \subseteq \bar{\sP} := \{1, \ldots, N\} \setminus \sP$ that will respectively keep track of the weights in $\sP$ that are no longer pruned and the weights in $\bar{\sP}$ that are pruned in lieu of those in $\sI$. 
For each weight $i \in \sP$, 
we calculate $\alpha_i$ in Line~\ref{lin:alpha} as the impact of pruning weight $i$ if the other pruned weights are also those in $\sP$. 
The weights in $\sP$ are then sorted in a sequence $\pi$ of nonincreasing $\alpha$ values in Line~\ref{lin:pi}, 
hence from largest to the smallest impact.
For each weight $j \in \bar{\sP}$, 
we calculate $\beta_j$ in Line~\ref{lin:beta} as the impact of having weight $j$ removed in addition to all the weights in $\sP$. 
The weights in $\bar{\sP}$ are then sorted in a sequence $\theta$ of nondecreasing impact if swapped with the first element $i := \pi_1$ in Line~\ref{lin:theta}. 
Hence, the first element $j := \theta_1$ yields the greatest reduction if swapped with $i$. 
If swapping $i$ and $j$ does not yield a reduction of at least $\varepsilon$, the local search stops in Line~\ref{lin:stop}. 
Otherwise, we loop with variable $i$ over the currently pruned weighs in $\sP$ and with variable $j$ over the unpruned weights in $\bar{\sP}$ between Line~\ref{lin:forij_begin} and Line~\ref{lin:forij_end}. 
The weights in $\sP$ are visited according to the sequence $\pi$, 
and for the element at the $ii$-th position of $\pi$ we consider only the elements in $\bar{\sP}$ between positions $ii - \rho$ and $ii + \rho$ of sequence $\theta$. 
There is no guarantee that the subsequent pairs of weights are sorted from largest to smallest reduction if swapped, 
but we loop on those to amortize the large cost of sorting the parameters in sequences $\pi$ and~$\theta$. 
We keep the number of steps approximately linear by limiting that each element in $\sP$ can only be swapped with at most $2 \rho + 1$ elements in $\bar{\sP}$. 
In addition, the loop is interrupted once the number of parameters in $\sP$ that could not be swapped reaches $\tau$. 
Parameter $\varepsilon$ is used in Line~\ref{lin:eps1} and Line~\ref{lin:eps2} to restrict changes to the pruning set $\sP$ to cases in which the local estimate of the loss function improves by at least that much. Otherwise, we can observe that after a few swapping only minor changes on the loss occur and thus one weight may swap in and out of $\sP$ again. 
To simplify notation, we do not divide all the calculated impacts by 2. For a concrete illustration, we visualize the basic swapping operation in Figure~\ref{fig:vis-cbs-s}.


We refer to Appendix~\ref{ap:matrix_ops} for the matrix operations to calculate $\alpha$, $\beta$, and $\gamma$ with GPUs; 
and we refer 
to Appendix~\ref{ap:hessian} for
how to  approximate and decompose the Hessian matrix $\mH$ in order to reduce time and space complexity.

\begin{algorithm}[h!]
   \caption{Prune Selection Swapping Local Search}
   \label{alg:local_search}
\begin{algorithmic}[1]
   \STATE {\bfseries Input:} 
   $\sP$ - initial set of pruned weights, 
   $\varepsilon$ - min impact variation for changing $\sP$, 
   $\tau$ - max failed weight swap attempts in $\sP$, 
   $\rho$ - range of candidates for each weight swap, 
   $steps_{max}$ - max number of steps, 
   $noimp_{max}$ - max number of non-improving steps 
   \STATE {\bfseries Output:} updated set of pruned weights $\sP_F$
   \STATE Initialize $\sP_F \gets \sP, sImprove \gets 0$,
   \FOR{$s \gets 1$ {\bfseries to} $steps_{max}$}
   \STATE $\sI \gets \emptyset; \sJ \gets \emptyset$
   \FOR{$i \in \sP$}
   \STATE $\alpha_i \gets \bar{\evw}_i \mH_{i,i} \bar{\evw}_i$ \\  \qquad ~ ~ $+ \sum_{j \in \sP : j \neq i} \left( \bar{\evw}_i \mH_{i,j} \bar{\evw}_j + \bar{\evw}_j \mH_{j,i} \bar{\evw}_i \right)$ \alglinelabel{lin:alpha}
   \ENDFOR
   \STATE Compute a sequence $\pi$ of $i \in \sP$ by nonincreasing $\alpha_i$ 
   \alglinelabel{lin:pi}
   \FOR{$j \in \bar{\sP} := \{1, \ldots, N\} \setminus \sP$}
   \STATE $\beta_j \gets \bar{\vw}_j \mH_{j,j} \bar{\vw}_j$ \\  \qquad ~ ~ $+ \sum_{i \in \sP} \left( \bar{\vw}_i \mH_{i,j} \bar{\vw}_j + \bar{\vw}_j \mH_{j,i} \bar{\vw}_i \right)$ \alglinelabel{lin:beta}
   \ENDFOR
   \STATE Compute a sequence $\theta$ of $j \in \bar{\sP}$ by nondecreasing $\beta_j - \gamma_{{\pi_1},j}$, 
   where $\gamma_{i,j} := \bar{\evw}_i \mH_{i,j} \bar{\evw}_j + \bar{\evw}_j \mH_{j,i} \bar{\evw}_i$
   \alglinelabel{lin:theta}
   \IF{$\beta_{\theta_1} - \gamma_{{\pi_1},{\theta_1}} - \alpha_{\pi_1} > - \varepsilon$} \alglinelabel{lin:eps1}
   \STATE {\bfseries Terminate} \alglinelabel{lin:stop}
   \ENDIF
   \STATE $c=0$
   \FOR{$i \in [\pi_1, \ldots, \pi_{|\sP|}]$} \alglinelabel{lin:forij_begin}
   \STATE $ii \gets$ index of $i$ in $\pi$; $c\gets c+1$
   \FOR{$j  \in [\theta_{\max\{1, ii-\rho\}}, \ldots, \theta_{\min\{|\bar{\sP}|, ii+\rho\}}] \setminus \sJ$}
   \IF{$\left(\beta_j + \sum\limits_{j' \in \sJ} \gamma_{j,j'}  - \sum\limits_{i' \in \sI} \gamma_{i',j} \right) - \left(  \gamma_{i,j} \right) - \left( \alpha_i + \sum\limits_{j' \in \sJ} \gamma_{i,j'} - \sum\limits_{i' \in \sI} \gamma_{i,i'} \right)  \leq - \varepsilon$} \alglinelabel{lin:eps2}
   \STATE $\sI \gets \sI + i$; $\sJ \gets \sJ + j$; $c \gets c-1$
   \STATE {\bfseries Goto Line 26}
   \ENDIF
   \ENDFOR
   \IF{$c \geq \tau$}
   \STATE {\bfseries Goto Line 30}
   \ENDIF
   \ENDFOR \alglinelabel{lin:forij_end}
   \IF{$\sI = \emptyset$}
   \STATE {\bfseries Terminate}
   \ENDIF
   \STATE $\sP \gets \sP \cup \sJ \setminus \sI$
   \IF{$\mathcal{L}(\sP) < \mathcal{L}(\sP_F)$}
   \STATE $\sP_F \gets \sP$; $sImprov \gets s$
   \ELSIF{$(s - sImprov) > noimp_{max}$}
   \STATE {\bfseries Terminate}
   \ENDIF
   \ENDFOR
\end{algorithmic}
\end{algorithm}

\section{A Greedy Randomized Constructive Algorithm for Pruning Selection}\label{sec:rmp}

The success of local search typically depends on the quality of the initial solution. 
Since magnitude-based pruning can be rather effective as a stating point, even if not always perfect,  
we consider how to leverage its guidance while applying the diversification tricks of old-school optimizers. 

In particular, 
we propose a constructive heuristic algorithm along the lines of semi-greedy methods~ \citep{hart1987semigreedy,feo1989setcovering,feo1995grasp}. These methods rely on a metric for identifying elements that are generally associated with good solutions. 
In our case, 
we know that selecting the weights with the smallest absolute value tends to be effective. 
In addition, 
some randomness is introduced in the actual selection of elements by semi-greedy methods, 
by which we may not necessarily always obtain a solution with the top-ranked elements for the chosen metric. 
By repeating the construction a sufficient number of times, 
we sample several solutions that closely follow the metric rather than obtain one solution that follows it rather strictly.

\cref{alg:constructive} describes our constructive method. 
The outer loop from Line~\ref{lin:outer_for_begin} to Line~\ref{lin:inner_for_begin} produces $S$ selections of weights to prune, among which the one with smallest sample loss is chosen. 
The inner loop from Line~\ref{lin:inner_for_end} to Line~\ref{lin:outer_for_end} iteratively partitions the weights of the neural network into $B$ buckets of roughly the same size, 
and in which of those we select the same proportion of weights to be pruned. 
The partitioning step introduces randomness to the weight selection while still making the weights with smallest absolute value more likely to be pruned. 
This construction is highly parallelizable because the processing of each bucket can be done by a separate unit, 
hence favoring the use of modern GPUs.

\begin{algorithm}[h!]
   \caption{Greedy Randomized Pruning Selection}
   \label{alg:constructive}
\begin{algorithmic}[1]
   \STATE {\bfseries Input 1:} number of weight-partitioning buckets $B$
   \STATE {\bfseries Input 2:} number of pruning sets  $S$ to be generated
   \STATE {\bfseries Output:} best set of pruned weights $\sP$ found
   \FOR{$s \gets 1$ {\bfseries to} $S$} \alglinelabel{lin:outer_for_begin}
   \STATE $\sQ \gets \emptyset$
   \STATE $\sW \gets \{1, \ldots, N\}$
   \FOR{$k \gets 1$ {\bfseries to} $B$} \alglinelabel{lin:inner_for_begin}
   \STATE $\sB \gets$ Randomly pick $\min\{ \lceil \frac{N}{B} \rceil, |
   \sW| \}$ weights from $\sW$
   \STATE $\sW \gets \sW \setminus \sB$
   \STATE $\sB' \gets$ Magnitude-based pruning selection on $\sB$
   \STATE $\sQ \gets \sQ \cup \sB'$ 
   \ENDFOR 
   \alglinelabel{lin:inner_for_end}
   \IF{$s = 1$ {\bfseries or} sample loss is smaller pruning $\sQ$ than $\sP$}
   \STATE $\sP \gets \sQ$
   \ENDIF
   \ENDFOR \alglinelabel{lin:outer_for_end}
\end{algorithmic}
\end{algorithm}

\FloatBarrier

\section{The Systematic Weight Update for OBS with Multiple Pruned Weights}\label{sec:update}

We resort to the same weight update technique described for one pruned weight by \citet{hassibi1992surgeon} and for two pruned weights to illustrate generalization by \citet{singh2020woodfisher}. 
However, we take the selection of weights to be pruned as a given from CBS-S or some other pruning method. 
To facilitate the parallel, we follow the notation of prior work as close as possible, starting with the perturbation $\delta \vw = \vw - \bar{\vw}$ and $\ve_i$ is the $i^{\text{th}}$ canonical basis vector. 
We dualize the constraint of CBS-U with the Lagrange multiplier $\boldsymbol\lambda$ to obtain the following Lagrangian $L(\vw, \boldsymbol\lambda)$:
\[
        L(\delta \vw, \boldsymbol\lambda)  = \frac{1}{2} \delta \vw^{T} \mH \delta \vw + \sum_{i \in \sP} \lambda_i (\ve_i^T \delta \vw + \bar{\evw}_i) 
\]
We use the solution $\delta \vw^*$ of $\nabla L(\delta \vw, \lambda) = 0$ to obtain the Lagrange dual function $g(\lambda)$ for the infimum of $L(\delta \vw, \lambda)$ and then 
obtain the maximizer $\boldsymbol\lambda^*$ of $g(\boldsymbol\lambda)$ by solving the system $\nabla g (\boldsymbol\lambda) = 0$ (see Appendix \ref{ap:cbs-u} for derivation):
\begin{align*}
\nabla_{\lambda_i} g(\boldsymbol\lambda) = - \sum_{j \in \sP} \lambda^*_j e^T_i \mH^{-1} e_j + \bar{\evw}_i = 0 \qquad \forall i \in \sP
\end{align*}
Hence, if we denote by  $[\mH^{-1}]_{\sP, \sP}$ the submatrix of $\mH^{-1}$ on the rows and columns of the pruned weights $\sP$, 
then the updated vectors of weights $\vw^*$ minimizing the approximate loss function with the weights in $\sP$ pruned is the following:
\[
\vw^* = \boldsymbol\delta \vw^* - \bar{\vw} = - [\mH^{-1}]_{\sP, \sP} \boldsymbol\lambda^* - \bar{\vw}
\]

\section{Experimental Evaluation}\label{sec:experiments}

To validate our approach, we have applied it to compress commonly used networks for image classification. Given a well-trained model, we obtain sparse models and then we compare them with those obtained by both a state-of-the-art method as well as a commonly used method for unstructured pruning. Especially, we make detailed comparisons with WoodFisher~(WF)~\citep{singh2020woodfisher}, which is the state-of-the-art derived from OBS; including an adaption of WoodFisher, which we denote WoodFisher-S~(WF-S), by which the remaining weights are not pruned and thus we restrict ourselves to pruning selection. We also include results for Magnitude-based Pruning~(MP). We have included the Pytorch source code in Supplementary materials. 

\subsection{Experimental Setup}
Although powerful for network pruning, the second-order approximation is expensive to apply on large networks. We make our algorithms more efficient by using the empirical Fisher and WoodFisher approximations of $\mH$ and $\mH^{-1}$. 

\noindent\textbf{Empirical Fisher} As described in the methodology, our approach requires the Hessian $\mH$ and its inverse $\mH^{-1}$. 
In order to make them applicable to large networks, such as MobileNet ($\sim$ 6.1M parameters), we use the empirical Fisher matrix to approximate $\mH$.  With such a decomposition of $\mH$, if we randomly sample a subset of the training data of size $K$ and the model has $N$ parameters, the required space complexity is $O(K\times N)$ instead of $O(N^2)$. 
\citet{singh2020woodfisher} note that a few hundred samples are sufficient for estimates applied to network pruning, 
and we have confirmed that while developing this work. 
Furthermore, we can parallelize the tensor computation with such a decomposition and further speed up the local search algorithm. We refer 
to Appendix~\ref{ap:hessian} for more details.

\noindent\textbf{WoodFisher Approximation} Due to the runtime cubic in the number of model parameters,  it is inviable to directly compute $\mH^{-1}$. Hence, we adopt the WoodFisher approximation by \citet{singh2020woodfisher}, which is  an efficient method to estimate $\mH^{-1}$ in $\mathcal{O}(K\times K)$ time and can be further reduced with a block-wise approximation in large networks. We use the same number of samples and block sizes as \citet{singh2020woodfisher}.

\noindent\textbf{Pre-Trained Models} We present our results on the MOBILENETV1 model of the STR method~\citep{mobilenets} trained on ImageNet~\citep{imagenet}, ResNet20~\citep{he2016deep} and CIFARNet model~\citep{krizhevsky2009learning} (a revised AlexNet consisted of three convectional layers and two fully connected layers) trained on CIFAR10~\citep{krizhevsky2009learning}, and MLPNet of three linear layers ($3072\rightarrow{40}\rightarrow{20}\rightarrow{10}$) trained on MNIST~\citep{lecun1998mnist}. To make a fair comparison with WoodFisher, we use the same checkpoints provided in its open-source implementation.

\noindent\textbf{Local Search Parameters} We use the following parameters for the local search algorithm: $\varepsilon = 1e-4$, $\tau = 20$, 
$\rho = 10$, 
$steps_{max} = 50$, 
and $noimp_{max} = 5$. We order  $\alpha$ and $\beta$ at the beginning  of local search.
We have chosen the parameters based
on preliminary experiments with MLPNet to strike a balance between performance and speed.


\noindent\textbf{Sparsity Rates} We have initially experimented with sparsity rates ranging from 0.1 to 0.9 in multiples of 0.1. Having observed that the results across methods were very similar for lower rates, 
we have focused on higher  rates provided that the accuracy remained better than random guessing, e.g., more than 10\% accuracy for at least one model for uniformly distributed datasets with 10 classes each. 

We show the pruning performance in Tables \ref{table:mlpnet.rst}, \ref{table:resnet20.rst}, \ref{table:cifarnet.rst}, \ref{table:mobilenet.rst} and~\ref{table:ablation.update}. We focus on where the methods start to differ from the original models and do not report very low sparsity, since all the methods exhibit similar performance. In  Appendix~\ref{ap:more_exp}, we supplement with more results, including running time as well as ablation studies. 

\subsection{CBS-S Benchmarking}
We first compare our CBS-S approach, consisting of selection algorithms RMP and LS with the results obtained with MP as well as with WoodFisher selection (WF-S). 
These results consist of the leftmost part of Tables \ref{table:mlpnet.rst}, \ref{table:resnet20.rst}, \ref{table:cifarnet.rst}, and \ref{table:mobilenet.rst}. 
%
In general, CBS-S produces more accurate models than other selection methods. CBS-S alone even outperforms WoodFisher with weight update under extreme sparsity on a few tasks (see CBS-S and WF in Tables \ref{table:mlpnet.rst}, \ref{table:resnet20.rst}, \ref{table:cifarnet.rst}). 

\subsection{CBS Benchmarking}
Our next results compare our CBS approach, consisting of CBS-S as well as the systematic weight update for CBS-U, with the results obtained with WoodFisher. These results consist of the rightmost part of Tables \ref{table:mlpnet.rst}, \ref{table:resnet20.rst}, \ref{table:cifarnet.rst}, and \ref{table:mobilenet.rst}.
In most of the cases, CBS produces more accurate models than WF.

\subsection{CBS-U Benchmarking}

We also compare the benefit of CBS-U alone by applying it to the pruning selection of MP and WF-S in Table~\ref{table:ablation.update}.
Except for cases of very low sparsity, applying CBS-U to the pruned model improves its accuracy by a large margin. 
Meanwhile, we notice that CBS-U decrease performance a bit under extreme sparsity (greater than 0.9) on simple tasks (see CBS-S and CBS in Tables \ref{table:mlpnet.rst}, \ref{table:resnet20.rst}, \ref{table:cifarnet.rst}). 
We discuss this in Appendix~\ref{ap:more_exp} and hope to investigate in the future.

\begin{table*}[]
    \centering
    \begin{small}
        \begin{tabular}{@{\extracolsep{8pt}}cccccccc}
            & \multicolumn{4}{c}{Prune Selection} & \multicolumn{3}{c}{Weight Update} \\
            \cline{2-5} \cline{6-8}
Sparsity &  \textit{MP} &\textit{WF-S}   &\textit{CBS-S} &\textbf{Improvement}   &\textit{WF} &\textit{CBS} & \textbf{Improvement} \\
\cline{1-1}
\cline{2-2}
\cline{3-3}
\cline{4-4}
\cline{5-5}
\cline{6-6}
\cline{7-7}
\cline{8-8}

0.50 &   93.93   &93.92 &93.91  &-0.02  &\best{94.02} &\second{93.96} & -0.06\\
0.60 &   93.78   &93.748 &93.85 &0.07   &\second{93.82} &\best{93.96} & 0.14 \\
0.70 &   93.62   &93.48 &93.75 &0.13   &\second{93.77} &\best{93.98} &0.21 \\
0.80 &   92.89   &93.13 &\second{93.59}  &0.46   &93.57 &\best{93.90} &0.33 \\
0.90 &   90.30    &90.77 &\second{92.37} &1.60   &91.69  &\best{93.14} &1.45 \\
0.95&   83.64   &83.16 &\second{88.24} &4.60   &85.54 &\best{88.92} &3.38 \\
0.98&   32.25   &34.55  &\best{66.64} &32.09  &38.26 &\second{55.45} & 17.20 \\
            \bottomrule
        \end{tabular}
    \end{small}
    \vspace{-0.1in}
    \caption{The pruning performance of different methods on MLPNet trained on MNIST. The accuracy of the model before pruning is 93.97\%. The results were averaged over five runs. The \best{best} and \second{second} best results are highlighted.}
    \label{table:mlpnet.rst}
    \vspace{-0.1in}
\end{table*}

\begin{table*}[]
    \centering
    \begin{small}
        \begin{tabular}{@{\extracolsep{8pt}}cccccccc}
            & \multicolumn{4}{c}{Prune Selection} & \multicolumn{3}{c}{Weight Update} \\
            \cline{2-5} \cline{6-8}
Sparsity &  \textit{MP} &\textit{WF-S}   &\textit{CBS-S} &\textbf{Improvement}   &\textit{WF} &\textit{CBS} & \textbf{Improvement} \\
\cline{1-1}
\cline{2-2}
\cline{3-3}
\cline{4-4}
\cline{5-5}
\cline{6-6}
\cline{7-7}
\cline{8-8}

0.30&    90.77&  90.81& 90.97   &0.16 &  \best{91.37} &\second{91.35}&    -0.01 \\
0.40&    89.98&  90.02 & 90.66  &0.64 &  \second{91.15} &\best{91.21}&    0.05  \\
0.50&    88.44&  88.06& 89.32   &0.88 & \second{90.23} &\best{90.58}&    0.35  \\
0.60&    85.24&  84.95& 86.48  &1.24 &  \second{87.96} &\best{88.88} &    0.92  \\
0.70&    78.79&  78.09& 79.55  &0.76 &  \second{81.05} &\best{81.84}&    0.79  \\
0.80&    54.01&  52.05 & \second{61.30}    &7.29 &  \best{62.63} &51.28&    -11.35 \\
0.90&    11.79&  11.44& \best{16.83}  &5.04 &  11.49 & \second{13.68}&    2.19  \\
            \bottomrule
        \end{tabular}
    \end{small}
    \vspace{-0.1in}
    \caption{The pruning performance of different methods on ResNet20 trained on Cifar10. The accuracy of the model before pruning is 91.36\%. The results were averaged over five runs. The \best{best} and \second{second} best results are highlighted.}
    \label{table:resnet20.rst}
    \vspace{-0.1in}
\end{table*}

\begin{table*}[]
    \centering
    \begin{small}
        \begin{tabular}{@{\extracolsep{8pt}}cccccccc}
            & \multicolumn{4}{c}{Prune Selection} & \multicolumn{3}{c}{Weight Update} \\
            \cline{2-5} \cline{6-8}
Sparsity &  \textit{MP} &\textit{WF-S}   &\textit{CBS-S} &\textbf{Improvement}   &\textit{WF} &\textit{CBS} & \textbf{Improvement} \\
\cline{1-1}
\cline{2-2}
\cline{3-3}
\cline{4-4}
\cline{5-5}
\cline{6-6}
\cline{7-7}
\cline{8-8}

0.50  & 79.77&   79.76   & \best{79.84} &  0.08   &79.76& \second{79.82}  &0.06 \\
0.60  & \best{79.90} &   \second{79.87}   & 79.86 &  -0.01  &79.77& 79.85  &0.08 \\
0.70  & 79.49&   \best{79.78}   & 79.47 &  -0.31  &79.65& \second{79.72}  &0.07 \\
0.80  & 76.92&   77.24   & 77.82 &  0.58   &\best{78.72}& \second{78.58}  &-0.14 \\
0.90  & 49.91&   52.66   & \second{62.09} &  9.43   &61.95& \best{65.52}  &3.57 \\
0.95 & 17.61&   14.81   & \best{32.63}&  17.82  &15.29& \second{21.07}  &5.78 \\
0.98 & 10.12&   9.34    & \best{15.49} &  6.15   &9.90  &  \second{12.16} &2.26 \\

            \bottomrule
        \end{tabular}
    \end{small}
    \vspace{-0.1in}
    \caption{The pruning performance  of different methods on CIFARNet trained on Cifar10. The accuracy of the model before pruning is 79.75\%. The results were averaged over five runs. The \best{best} and \second{second} best results are highlighted.}
    \label{table:cifarnet.rst}
    \vspace{-0.1in}
\end{table*}

\begin{table*}[]
    \centering
    \begin{small}
        \begin{tabular}{@{\extracolsep{8pt}}cccccccc}
            & \multicolumn{4}{c}{Prune Selection} & \multicolumn{3}{c}{Weight Update} \\
            \cline{2-5} \cline{6-8}
Sparsity &  \textit{MP} &\textit{WF-S}   &\textit{CBS-S} &\textbf{Improvement}   &\textit{WF} &\textit{CBS} & \textbf{Improvement} \\
\cline{1-1}
\cline{2-2}
\cline{3-3}
\cline{4-4}
\cline{5-5}
\cline{6-6}
\cline{7-7}
\cline{8-8}
0.30&    71.60& 71.68   &71.48& -0.20    &\best{71.88}  &\best{71.88}& 0.00     \\
0.40&    69.16& 69.53   &69.37& -0.16   &\second{71.15}  &\best{71.45}& 0.30   \\
0.50&    62.61 &  63.10   &62.96& -0.14   &\second{68.91}  &\best{70.21}& 1.30   \\
0.60&    41.94 &  44.07  &43.10 &  -0.97  &\second{60.90}   &\best{66.37}& 5.47  \\
0.70&    6.78 &  7.33   &7.63 &  0.3    &\second{29.36}  &\best{55.11}& 25.75  \\
0.80&    0.11 &  0.11   &0.12 &  0.01   &\second{0.24}   &\best{16.38}& 16.14 \\

            \bottomrule
        \end{tabular}
    \end{small}
    \vspace{-0.1in}
    \caption{The pruning performance of different methods on MobileNet trained on ImageNet. The accuracy of the model before pruning is 72.0\%. The \best{best} and \second{second} best results are highlighted.}
    \label{table:mobilenet.rst}
    \vspace{-0.1in}
\end{table*}

\begin{table*}[]
    \centering
    \begin{small}
        \begin{tabular}{@{\extracolsep{8pt}}ccccccc}
            & \multicolumn{3}{c}{Magnitude-Based Pruning} & \multicolumn{3}{c}{WoodFisher} \\
            \cline{2-4}
            \cline{5-7}
            Sparsity & \textit{MP} & \textit{MP + CBS-U} & \textbf{Improvement} & \textit{WF} & \textit{WF-S + CBS-U} & \textbf{Improvement} \\
\cline{1-1}
\cline{2-2}
\cline{3-3}
\cline{4-4}
\cline{5-5}
\cline{6-6}
\cline{7-7}
0.30&    71.61 & \best{71.88}   & 0.27   & \best{71.88}   & 71.86 & -0.02 \\
0.40&    69.16 & \best{71.43}   & 2.27   & 71.15   & \best{71.43} & 0.28 \\
0.50&    62.61 &  \second{70.24}  & 7.63   & 68.91   & \best{70.30} & 1.39   \\
0.60&    41.94 &  66.33  & 24.39  & \best{60.90}   & \second{66.62} & 5.72 \\
0.70&    6.78  &  \second{55.51}  & 48.73  & 29.36   & \best{56.09} & 26.73 \\
0.80&    0.11  &  \second{16.58}  & 16.47  & 0.24    & \best{17.94} & 17.70 \\
            \bottomrule
        \end{tabular}
    \end{small}
    \vspace{-0.1in}
    \caption{Ablation study on CBS-U and comparision with the weight update of WoodFisher method. This is tested on MobileNet trained on ImageNet with the accuracy of 72.0\% before pruning. The \best{best} and \second{second} best results are highlighted.}
    \label{table:ablation.update}
    \vspace{-0.1in}
\end{table*}










\section{Conclusion}\label{sec:conclusion}

\epigraph{\textit{Optimization matters only when it matters. When it matters, it matters a lot, but until you know that it matters, don't waste a lot of time doing it.}}{\citet{flounder1996}}

We have introduced the Combinatorial Brain Surgeon~(CBS), an approach to network pruning that accounts for the joint effect of pruning multiple weights of a neural network. 
Our approach is based on the same paradigm as the classic Optimal Brain Surgeon~(OBS), 
in which the selection of pruned weights is also followed by the update of remaining weights and both aim to minimize a local quadratic approximation of the sample loss. 
We obtain a tractable approach by  
dissociating pruning selection and  weight update as distinct optimization problems: 
CBS Selection~(CBS-S) and CBS Update~(CBS-U), respectively. 

One benefit of this dissociation is that CBS Selection can be used as a stand-alone pruning technique if updating the remaining weights is not in the scope of the pruning algorithm, such as when pruning is performed at initialization or during training. 
Due to the presently large number of parameters in neural networks, 
CBS-S remains a challenging problem for a straightforward approach with modern optimization solvers. 
Nevertheless, 
we resort to tried-and-tested techniques that were once the best resort to tackle many discrete optimization problems:  
we design a local search method intended to leverage the synergy between pruned weights to minimize the sample loss approximation; and we design a semi-greedy heuristic to leverage the effectiveness of magnitude-based pruning. 

We have observed competitive results for our approach to CBS-S as well as to CBS by also extending the weight update from OBS to multiple parameters with CBS-U. 
More specifically, 
we obtain substantially more accurate models in both cases for higher sparsity rates, such as 0.90, 0.95, and 0.98. 
We believe that this is particularly encouraging, 
since the purpose of using network pruning is to reduce the number of parameters by the largest possible amount.

To the best of our understanding, 
this is the first paper that considers joint impact in network pruning at the parameter level. 
That comes with a rather challenging optimization problem, 
to which we provide a first approach. 
We hope to see it further improved in future work, as we believe that this is genuinely a case in which optimization matters a lot!

\vspace{-0.1in}
\section*{Acknowledgement}
\vspace{-0.1in}
This project has been supported by NSF IIS-1764071, NSF IIS-1910983 and NSF CAREER Award IIS-2046295. Thiago Serra was supported by NSF IIS-2104583. We thank the anonymous reviewers for their valuable and constructive feedback that helped in shaping the final manuscript.

\FloatBarrier








\bibliography{references}
\bibliographystyle{icml2022}

\newpage
\appendix
\onecolumn

 \section{Solving the Systematic Weight Update for Multiple Pruned Weights}\label{ap:cbs-u}
 We resort to the same weight update technique described for one pruned weight by \citet{hassibi1992surgeon} and for two pruned weights to illustrate generalization by \citet{singh2020woodfisher}. 
However, we take the selection of weights to be pruned as a given from CBS-S or some other pruning method. 
To facilitate the parallel, we follow the notation of prior work as close as possible, starting with the perturbation $\delta \vw = \vw - \bar{\vw}$ and $\ve_i$ is the $i^{\text{th}}$ canonical basis vector. 
We dualize the constraint of CBS-U with the Lagrange multiplier $\boldsymbol\lambda$ to obtain the following Lagrangian $L(\vw, \boldsymbol\lambda)$:
\[
        L(\delta \vw, \boldsymbol\lambda)  = \frac{1}{2} \delta \vw^{T} \mH \delta \vw + \sum_{i \in \sP} \lambda_i (\ve_i^T \delta \vw + \bar{\evw}_i) 
\]
We use the solution $\delta \vw^*$ of $\nabla L(\delta \vw, \lambda) = 0$ to obtain the Lagrange dual function $g(\lambda)$ for the infimum of $L(\delta \vw, \lambda)$ and then 
obtain the maximizer $\boldsymbol\lambda^*$ of $g(\boldsymbol\lambda)$ by solving the system $\nabla g (\boldsymbol\lambda) = 0$ :
\begin{align*}
\nabla L(\vw, \boldsymbol\lambda) = \mH  \delta \vw^* + \sum_{i \in \sP} \lambda_i \ve_i = 0 \\ ~ ~ \rightarrow \delta \vw^* = - \sum_{i \in \sP} \lambda_i \mH^{-1} \ve_i
\end{align*}
\begin{align*}
g(\boldsymbol\lambda) & = \frac{1}{2} \left( \sum_{i \in \sP} \lambda_i \mH^{-1} \ve_i \right)^T \mH \left( \sum_{i \in \sP} \lambda_i \mH^{-1} \ve_i \right) \\
& ~ ~ + \sum_{i \in \sP} \lambda_i \left( - e_i^T \sum_{j \in \sP} \lambda_j H^{-1} e_j + \bar{\evw}_i \right) \\
& = \frac{1}{2} \sum_{i \in \sP} \sum_{j \in \sP} \lambda_i \lambda_j e_i^T H^{-1} e_j \\ & ~ ~ - \sum_{i \in \sP} \sum_{j \in \sP} \lambda_i \lambda_j e_i^T H^{-1} e_j + \sum_{i\in \sP} \lambda_i \bar{\evw}_i \\ 
& = - \frac{1}{2} \sum_{i \in \sP} \sum_{j \in \sP} \lambda_i \lambda_j e_i^T H^{-1} e_j + \sum_{i \in \sP} \lambda_i \bar{\evw}_i 
\end{align*}

We then obtain the maximizer $\boldsymbol\lambda^*$ of the Lagrange dual function $g(\boldsymbol\lambda)$ by solving the system $\nabla g (\boldsymbol\lambda) = 0$:

\begin{align*}
\nabla_{\lambda_i} g(\boldsymbol\lambda) = - \sum_{j \in \sP} \lambda^*_j e^T_i \mH^{-1} e_j + \bar{\evw}_i = 0 \qquad \forall i \in \sP
\end{align*}

Hence, if we denote by  $[\mH^{-1}]_{\sP, \sP}$ the submatrix of $\mH^{-1}$ on the rows and columns of the pruned weights $\sP$, 
then the updated vectors of weights $\vw^*$ minimizing the approximate loss function with the weights in $\sP$ pruned is the following:

\[
\vw^* = \boldsymbol\delta \vw^* - \bar{\vw} = - [\mH^{-1}]_{\sP, \sP} \boldsymbol\lambda^* - \bar{\vw}
\]

\section{Approximating the Hessian and Matrix Operations for Local Search}\label{ap:matrix_ops}

If we denote the matrix of all network weights by $\vw$ and we index both $\vw$ and $\mH$ with subsets to denote submatrices, 
we can  calculate parameters $\boldsymbol\alpha$, $\boldsymbol\beta$, and $\boldsymbol\gamma$ through matrix operations that leverage the parallel processing of GPUs ($\odot$ is the element-wise product): 
\begin{align}
    \boldsymbol\alpha & = \vw_{\sP} \odot \left(\mH_{\sP,\sP} \vw_{\sP} \right) + \left(\vw_{\sP}^T \mH_{\sP,\sP}\right)^T \odot \vw_{\sP} 
    - \vw_{\sP} \odot \textit{diag}(\mH_{\sP,\sP}) \odot \vw_{\sP} \label{eq:chunk_alpha}\\
        \boldsymbol\beta & = \vw_{\overline{\sP}} \odot \left(\mH_{\overline{\sP},\sP} \vw_{\sP} \right) + \left({\vw_{\sP}}^T \mH_{\sP,\overline{\sP}}\right)^T \odot \vw_{\overline{\sP}} 
        +
\vw_{\overline{\sP}} \odot \textit{diag}(\mH_{\overline{\sP},\overline{\sP}}) \odot 
\vw_{\overline{\sP}} \label{eq:chunk_beta}\\
\boldsymbol\gamma_{\sP,\bar{\sP}} & = \vw_{\sP} \odot \left(\mH_{\sP, \bar{\sP}} \vw_{\bar{\sP}} \right) + \left(\vw_{\bar{\sP}}^T \mH_{\bar{\sP},\sP}\right)^T \odot \vw_{\sP} \label{eq:chunk_gamma}
\end{align}

\section{Approximating and Decomposing $\mH$ to Reduce Time and Space Complexity}\label{ap:hessian}

As we know, $\mH\in \mathcal{R}^{\mathcal{D} \times \mathcal{D}}$ will consume large memory, especially for large network, when calculating $\boldsymbol\alpha, \boldsymbol\beta,$ and  $\boldsymbol\gamma$. We will alleviate this by the following formulation.

\noindent\textbf{Fisher Matrix} In this method, we approach the Hessian matrix with the Fisher matrix. Please note that one set of gradients can be from a single sample or be average over a batch of samples, and $\mathcal{N}$ is the number of gradient sets.  
\[
H \approx F = \frac{1}{\mathcal{N}} \sum_{n=1}^{\mathcal{N}}\nabla \mathcal{L}_n \nabla \mathcal{L}_n^T,\quad \nabla \mathcal{L}_n  = \nabla \mathcal{L}\left(y_n, f\left(x_n;W\right) \right)
\]
Let's describe $\nabla \mathcal{L}_n $ as $g^n \in \mathcal{R}^{\mathcal{D}}$ and have $H\approx \frac{1}{\mathcal{N}}\sum_{n=1}^{\mathcal{N}} g^n \left(g^n\right)^T$. 

\noindent\textbf{Chunk operations} As shown in Equations~\ref{eq:chunk_alpha},~\ref{eq:chunk_beta},~\ref{eq:chunk_gamma}, the common operation for $\alpha, \beta, \gamma$ are the multiplication on the sub-matrix of $\mH$ and the sub-vector of $\vw$. Meanwhile, for efficiency, we split the layer-wise blocks of the Hessian matrix into smaller chunks along the diagonal as WoodFisher, which are smaller sub-matrix of $\mH$. Without loss of generality,  $\mH_{\sS_1\sS_2}$ indicates the sub-matrix of $\mH$ of row $\sS_1$ and columns $\sS_2$, $g_{\sS}$ and $\vw_{\sS}$ indicate the subset of gradients and weights respectively. We reformulate multiplication on the sub-matrix of $\mH$ and the sub-vector of $\vw$ as follows,
\begin{align*}
    \mH_{\sS_1\sS_2} \mathbf{w}_{\sS_2} & = \frac{1}{\mathcal{N}}\sum_{n=1}^{\mathcal{N}} g_{\sS_1}^n \left( g_{\sS_2}^n \right)^T \vw_{\sS_2} = \frac{1}{\mathcal{N}}\sum_{n=1}^{\mathcal{N}} g_{\sS_1}^n \left( \vw_{\sS2}^T g_{\sS_2}^n \right)^T \\
    \vw_{\sS_1}^T \mH_{\sS_1\sS_2} & = \frac{1}{\mathcal{N}}\sum_{n=1}^{\mathcal{N}} \vw_{\sS_1}^T g_{\sS_1}^n \left( g_{\sS_2}^n \right)^T 
\end{align*}

The computation cost for $ \mH_{\sS_1\sS_2} \vw_{\sS_2}$ is $|\sS_1| \times |\sS_2| \times |\sS_2|$ while it is $\mathcal{N} \times |\sS_1| + |\sS_2|\times |\sS_2|$ after decomposing $H$. Furthermore, the first one needs $\mathcal{D} \times \mathcal{D} + \mathcal{D}$ memory and the later one  only hold the gradients and the weights in memory with the memory complexity of $\mathcal{N} \times \mathcal{D} + \mathcal{D}$. 
With the above formulation, we can further reduce the FLOP and memory consumption for this algorithm. We can further parallize the later one to further speed it up.

\noindent\textbf{Hyper-parameters} 
As shown in Table~\ref{tab:fisher-param}, we follow the exact hyper-parameters as WoodFisher to estimate the Hessian matrix $\mH$ and its inverse$\mH^{-1}$.
To approximate $\mH$, we run the models on a few batches of the training samples to get their gradients. The number of batches(\textit{Subsample Size}) and the batch size (\textit{Mini-batch Size}) on different models are shown in the column of \textit{Fisher} of Table~\ref{tab:fisher-param}. When calculating the $\mH^{-1}$ on large models efficiently, WoodFisher split the layer-wise blocks of the Hessian matrix into smaller chunks along the diagonal (see \textit{Chunk Size} in Table~\ref{tab:fisher-param}). Our chunk operations above also use the same chunk size. Finally, we give out the batch size used in training the original models and evaluating the pruned models in the last column of Table~\ref{tab:fisher-param}.
\begin{table}[!h]
\centering
\setlength{\tabcolsep}{1.5pt}
{\small
\begin{tabular}{lcccc}
  \toprule
  \multirow{2}{*}{Model} & \multicolumn{2}{c}{Fisher} & Chunk & Batch \\ \cmidrule(lr){2-3}
                         & Subsample Size($\mathcal~{N}$)~             & Mini-batch Size             & Size & Size \\\midrule
 MLPNet   & 1000  & 1 & -- & 64 \\
 CIFARNet &  1000 & 1 & -- & 64 \\
 ResNet20 & 1000  & 1 & -- & 64 \\
 MobileNet&  400 & 2400 & 10000 & 128 \\
\bottomrule
\end{tabular}}
\caption{\label{tbl:test} Parameters used for WoodFisher approximation of $\mH^{-1}$. We unitize no chunking on MLPNet, block-wise estimation with respect to the layers without further chunking on CIFARNet and ResNet20.}
\label{tab:fisher-param}
\end{table}

\section{More experimental results}
\label{ap:more_exp}
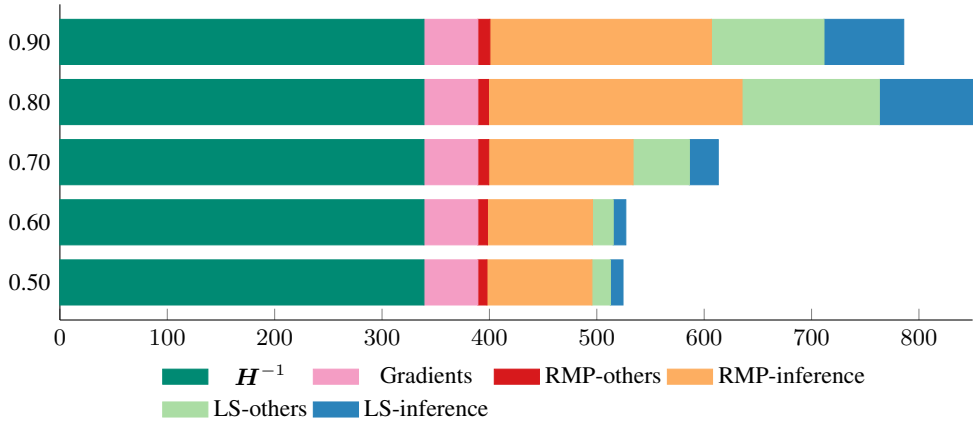
\begin{figure}
\centering
\begin{tikzpicture}
\begin{axis}[
    xbar stacked,
    legend style={
        legend columns=4,
        at={(xticklabel cs:0.5)},
        anchor=north,
        draw=none
    },
    ytick=data,
    axis y line*=none,
    axis x line*=bottom,
    tick label style={font=\footnotesize},
    legend style={font=\footnotesize},
    label style={font=\footnotesize},
    xtick={0,100,200,300,400,500,600,700,800},
    width=.8\textwidth,
    bar width=6mm,
    xlabel={Time in Seconds},
    yticklabels={0.50, 0.60, 0.70, 0.80, 0.90},
    xmin=0,
    xmax=850,
    area legend,
    y=8mm,
    enlarge y limits={abs=0.625},
]
\addplot[PineGreen,fill=PineGreen] coordinates
{(339, 0) (339, 1) (339, 2) (339, 3) (339, 4)};
\addplot[Lavender,fill=Lavender] coordinates
{(50.22, 0) (50.22, 1) (50.22, 2) (50.22, 3) (50.22, 4)};
\addplot[transfertoserver,fill=transfertoserver] coordinates
{(8.49,0) (8.92,1) (10.09,2) (9.87,3) (11.19,4)};
\addplot[database,fill=database] coordinates
{(97.71428571, 0) (97.79464286, 1) (134.4333333, 2) (236.1438356, 3) (206.3308824, 4)};
\addplot[transfertoclient,fill=transfertoclient] coordinates
{(17.24, 0) (19.34, 1) (52.47, 2) (127.88, 3) (104.81, 4)};
\addplot[rendering,fill=rendering] coordinates
{(11.72571429, 0) (11.73535714, 1) (26.88666667, 2) (108.6261644, 3) (74.27911765, 4)};
\legend{$\mH^{-1}$, Gradients, RMP-others,RMP-inference,LS-others,LS-inference}
\end{axis}
\end{tikzpicture}
\caption{Runing time of separate parts in CBS on CIFARNet}
\label{fig:runtime-ablation}
\end{figure}

\noindent\textbf{Running time} We have included plots comparing the running time of CBS and WoodFisher as shown in Figure 3. For high sparsity, in which weight update is less beneficial, we observe that the running time often makes CBS-S more advantageous because a simple approximation of the Hessian suffices for weight selection. We further study the running times of individual components of CBS as shown in Figure ~\ref{fig:runtime-ablation}. Both RMP and LS use the sample loss on a subset of training data (5000 samples on CIFARNet) as the criteria to pick up the best pruning set. We observe that this is time-consuming (see RMP-inference and LS-inference in Figure~\ref{fig:runtime-ablation}). The LS without inference (LS-others) is pretty fast. Recently, many subset selection methods~\cite{nguyen2021dataset, roh2021sample, ramalingam2021less} can get significantly smaller yet highly informative samples from the large training dataset. With this, we can do inference on a tiny dataset and thus greatly reduce our CBS-S running time. Please note that given $\mH^{-1}$, it spends little time to get the weight update for both CBS-U and WF-U. Thus we don't show it in Figure ~\ref{fig:runtime-ablation}.

\noindent\textbf{Ablation Study on CBS} We have included plots showing the average accuracy gain obtained with local search (\textbf{LS}), randomized magnitude pruning (\textbf{RMP}), and weight update (\textbf{Update}). We observe that (i) there is a small gain driven by weight updates for moderate sparsity (first and third plots in Figure 4); and (ii) a large gain driven by the heuristics for high sparsity (second and third plots); and (iii) the weight update based on CBS-S can decrease the performance under extreme sparsity. 
We believe that modifying the few remaining weights with weight update under high sparsity may sometimes change the model too much, especially since we start pruning large weights and thus push the Taylor approximation too far (as $\Delta \vw$ is too large). 
In WoodFisher, a scaling factor is used to control the impact of weight update. 
After applying the scaling factor, We have observed that weight update at sparsity 0.98 adds 0.5\% to accuracy rather than reducing it by 11.2\% as before (see current drop on second sub-figure of Figure 4).
Regarding the better performance of CBS-U on ImageNet as shown in Figure~\ref{table:mobilenet.rst}, we have used the same Hessian inverse approximations as WoodFisher and more samples are used for this dataset that for others(see Table~\ref{tab:fisher-param}). As shown in WoodFisher paper (table S2), using more samples for the approximation leads to better results. 

\noindent\textbf{Pruning performance Plots} Finally we visualize pruning performance of our CBS and the baselines from the Tables ~\ref{table:mlpnet.rst} and ~\ref{table:cifarnet.rst} as shown in Figure 5. 

\newcommand{\compressionPlotLegend}[9]{
    \begin{tikzpicture}[scale=0.43]
        \begin{axis}[legend pos=south west,xmin={#3},xmax={#4}, grid=major,
            xtick={#5},
            ytick={#6},
            xlabel={\huge{Sparsity}},
            ylabel={\huge{Test accuracy (\%)}},
            ticklabel style={font=\huge}]
            \addplot+[black, 
                line width={#7},
                scatter,
                mark=circle]
            table[x=sparsity,y=mp]
            {#1};
            \addlegendentry{\LARGE MP}
            \addplot+[color=blue, 
                line width={#7},
                dashed,
                scatter,
                mark=circle]
            table[x=sparsity,y=wf-s]
            {#1};
            \addlegendentry{\LARGE WF-S}
            \addplot+[color=blue, 
                line width={#7},
                scatter,
                mark=circle]
            table[x=sparsity,y=wf]
            {#1};
            \addlegendentry{\LARGE WF}
            \addplot+[color=red, 
                line width={#7},
                dashed,
                scatter,
                mark=circle]
            table[x=sparsity,y=cbs-s]
            {#1};
            \addlegendentry{\LARGE CBS-S}
            \addplot+[color=red, 
                line width={#7},
                scatter,
                mark=circle]
            table[x=sparsity,y=cbs]
            {#1};
            \addlegendentry{\LARGE CBS}
        \end{axis}
        \node at (#8,#9) {\small #2};
    \end{tikzpicture}
}

\newcommand{\compressionPlot}[9]{
    \begin{tikzpicture}[scale=0.43]
        \begin{axis}[xmin={#3},xmax={#4}, grid=major,
            xtick={#5},
            ytick={#6},
            xlabel={\huge{Sparsity}},
            ylabel={\huge{Test accuracy (\%)}},
            ticklabel style={font=\huge}]
            \addplot+[black, 
                line width={#7},
                scatter,
                mark=circle]
            table[x=sparsity,y=mp]
            {#1};
            \addplot+[blue, 
                line width={#7},
                dashed,
                scatter,
                mark=circle]
            table[x=sparsity,y=wf-s]
            {#1};
            \addplot+[blue, 
                line width={#7},
                scatter,
                mark=circle]
            table[x=sparsity,y=wf]
            {#1};
            \addplot+[red, 
                line width={#7},
                dashed,
                scatter,
                mark=circle]
            table[x=sparsity,y=cbs-s]
            {#1};
            \addplot+[red, 
                line width={#7},
                scatter,
                mark=circle]
            table[x=sparsity,y=cbs]
            {#1};
        \end{axis}
        \node at (#8,#9) {\small #2};
    \end{tikzpicture}
}

\newcommand{\ablationLegend}[9]{
    \begin{tikzpicture}[scale=0.43]
        \begin{axis}[legend pos=north west,xmin={#3},xmax={#4},grid=major,
            xtick={#5},
            y label style={above=-2mm},
            xlabel={\huge{Sparsity}},
            ylabel={\huge{Mean Acc. Gain}},
            ticklabel style={font=\huge}]
            \addplot+[blue, 
                line width={#7},
                scatter,
                mark=circle]
            table[x=sparsity,y=ls]
            {#1};
            \addlegendentry{\LARGE LS}
            \addplot+[blue, 
                line width={#7},
                dashed,
                scatter,
                mark=circle]
            table[x=sparsity,y=rmp]
            {#1};
            \addlegendentry{\LARGE RMP}
            \addplot+[red, 
                line width={#7},
                scatter,
                mark=circle]
            table[x=sparsity,y=update]
            {#1};
            \addlegendentry{\LARGE Update}
        \end{axis}
        \node at (#8,#9) {\small #2};
    \end{tikzpicture}
}

\newcommand{\ablationLegendRight}[9]{
    \begin{tikzpicture}[scale=0.43]
        \begin{axis}[legend pos=north west,xmin={#3},xmax={#4},
            xtick={#5},
            y label style={above=-2mm},
            xlabel={\huge{Sparsity}},
            ylabel={\huge{Mean Acc. Gain}},
            ticklabel style={font=\huge}]
            \addplot+[bblue, 
                line width={#7},
                scatter,
                mark=circle]
            table[x=sparsity,y=ls]
            {#1};
            \addlegendentry{\LARGE LS}
            \addplot+[borange, 
                line width={#7},
                dashed,
                scatter,
                mark=circle]
            table[x=sparsity,y=rmp]
            {#1};
            \addlegendentry{\LARGE RMP}
            \addplot+[borange, 
                line width={#7},
                scatter,
                mark=circle]
            table[x=sparsity,y=update]
            {#1};
            \addlegendentry{\LARGE Update}
            \addplot+[green, 
                line width={#7},
                scatter,
                mark=circle]
            table[x=sparsity,y=new_update]
            {#1};
            \addlegendentry{\LARGE 0.4 Update}
        \end{axis}
        \node at (#8,#9) {\small #2};
    \end{tikzpicture}
}

\newcommand{\runtime}[9]{
    \begin{tikzpicture}[scale=0.43]
        \begin{semilogyaxis}[xmin={#3},xmax={#4},
            xtick={#5},
            ytick={#6},
            yticklabels={#6},
            y label style={above=4mm},
            xlabel={\huge{Sparsity}},
            ylabel={\huge{Runtime (s)}},
            ticklabel style={font=\huge}]
            \addplot+[bmblue, 
                line width={#7},
                scatter,
                mark=circle]
            table[x=sparsity,y=wf]
            {#1};
            \addplot+[borange, 
                line width={#7},
                dashed,
                scatter,
                mark=circle]
            table[x=sparsity,y=cbs-s]
            {#1};
            \addplot+[borange, 
                line width={#7},
                scatter,
                mark=circle]
            table[x=sparsity,y=cbs]
            {#1};
        \end{semilogyaxis}
        \node at (#8,#9) {\small #2};
    \end{tikzpicture}
}

\newcommand{\runtimeLegend}[9]{
    \begin{tikzpicture}[scale=0.43]
        \begin{semilogyaxis}[xmin={#3},xmax={#4},
            legend style={at={(0.1,0.35)},anchor=south west},
            xtick={#5},
            ytick={#6},
            yticklabels={#6},
            y label style={above=4mm},
            xlabel={\huge{Sparsity}},
            ylabel={\huge{Runtime (s)}},
            ticklabel style={font=\huge}]
            \addplot+[bmblue, 
                line width={#7},
                scatter,
                mark=circle]
            table[x=sparsity,y=wf]
            {#1};
            \addlegendentry{\LARGE WF}
            \addplot+[borange, 
                line width={#7},
                dashed,
                scatter,
                mark=circle]
            table[x=sparsity,y=cbs-s]
            {#1};
            \addlegendentry{\LARGE CBS-S}
            \addplot+[borange, 
                line width={#7},
                scatter,
                mark=circle]
            table[x=sparsity,y=cbs]
            {#1};
            \addlegendentry{\LARGE CBS}
        \end{semilogyaxis}
        \node at (#8,#9) {\small #2};
    \end{tikzpicture}
}

\runtimeLegend{runtime_mlpnet.txt}{MLPNet}{0.5}{0.98}{0.5,0.6,0.7,0.8,0.9}{200,600,1600}{2.5pt}{2}{1}
\runtime{runtime_cifarnet.txt}{CIFARNet}{0.5}{0.98}{0.5,0.6,0.7,0.8,0.9}{200,400,800}{2.5pt}{2}{1}
\runtime{runtime_resnet20.txt}{ResNet20}{0.5}{0.98}{0.5,0.6,0.7,0.8,0.9}{200,400,800}{2.5pt}{2}{1}
\centerline{\textit{Figure 3. Runtime by sparsity for three models}. \label{fig:runtime}}

~ 

\ablationLegend{ablation_mlpnet.txt}{}{0.5}{0.8}{0.5,0.6,0.7,0.8}{92, 93, 94}{2.5pt}{2}{1}
\ablationLegend{ablation_mlpnet.txt}{MLPNet}{0.9}{0.98}{0.90, 0.95, 0.98}{92, 93, 94}{2.5pt}{2}{1}
\ablationLegend{ablation_cifarnet.txt}{CIFARNet}{0.5}{0.8}{0.5,0.6,0.7,0.8}{92, 93, 94}{2.5pt}{2}{1}
\ablationLegend{ablation_cifarnet.txt}{}{0.9}{0.98}{0.90, 0.95, 0.98}{92, 93, 94}{2.5pt}{2}{1} 
\centerline{\textit{Figure 4. Accuracy gain per step in MLPNet (leftmost two) and CIFARNet (rightmost two)}.}

\compressionPlotLegend{chart_mlpnet.txt}{}{0.5}{0.8}{0.5,0.6,0.7,0.8}{92, 93, 94}{2.5pt}{2}{1}
\compressionPlot{chart_mlpnet.txt}{MLPNet}{0.90}{0.98}{0.9, 0.95, 0.98}{30, 60, 90}{2.5pt}{2}{1}
\compressionPlot{chart_cifarnet.txt}{CIFARNet}{0.5}{0.8}{0.5,0.6,0.7,0.8}{76, 78, 80}{2.5pt}{2}{1}
\compressionPlot{chart_cifarnet.txt}{}{0.90}{0.98}{0.9, 0.95, 0.98}{0, 30, 60}{2.5pt}{2}{1}
\centerline{\textit{Figure 5. Accuracy change due to pruning in MLPNet (leftmost two) and CIFARNet(rightmost two)}.}


\end{document}